\documentclass{aircc}
\usepackage{mathpartir}
\usepackage{hyperref}
\usepackage{listings}
\usepackage{paralist}
\usepackage{stmaryrd}
\usepackage{graphicx}
\usepackage{algorithm2e}
\usepackage{amsmath}
\usepackage{adjustbox}

\begin{document}

\title{
Referring Expressions with Rational Speech Act Framework:\\ A Probabilistic Approach
}

\author{Hieu Le$^{1}$, Taufiq Daryanto$^{2}$, Fabian Zhafransyah$^{2}$, Derry Wijaya$^{1}$, Elizabeth Coppock$^{1}$, Sang Chin$^{1}$}
\affiliation{${^1}$ Boston University , Boston, MA, USA\\ \email{\url{hle@bu.edu}} \\
$^{2}$Institut Teknologi Bandung, Bandung, Indonesia
}
\maketitle

\begin{abstract}
This paper focuses on a referring expression generation (REG) task in which the aim is to pick out an object in a complex visual scene. One common theoretical approach to this problem is to model the task as a two-agent cooperative scheme in which a `speaker' agent would generate the expression that best describes a targeted area and a `listener' agent would identify the target. Several recent REG systems have used deep learning approaches to represent the speaker/listener agents. The Rational Speech Act framework (RSA), a Bayesian approach to pragmatics that can predict human linguistic behavior quite accurately, has been shown to generate high quality and explainable expressions on toy datasets involving simple visual scenes. Its application to large scale problems, however, remains largely unexplored. This paper applies a combination of the probabilistic RSA framework and deep learning approaches to larger datasets involving complex visual scenes in a multi-step process with the aim of generating better-explained expressions. We carry out experiments on the RefCOCO and RefCOCO+ datasets and compare our approach with other end-to-end deep learning approaches as well as a variation of RSA to highlight our key contribution. Experimental results show that while achieving lower accuracy than SOTA deep learning methods, our approach outperforms similar RSA approach in human comprehension and has an advantage over end-to-end deep learning under limited data scenario. Lastly, we provide a detailed analysis on the expression generation process with concrete examples, thus providing a systematic view on error types and deficiencies in the generation process and identifying possible areas for future improvements.
\end{abstract}


\section{Introduction}
Presented with a scene involving two dogs, where one has a frisbee in its mouth, native speakers of English will effortlessly characterize the lucky dog as {\em the dog with the frisbee}. Computers are not so good at this yet. The task in question is called referring expression generation (REG).

A common approach to REG is modeling the problem as a two-agent system in which a speaker agent would generate an expression given some input and a listener agent would then evaluate the expression. This modeling method is widely applied, for example in \cite{MonroeLearningRSA}. 

In the last few years, many attempts at REG have applied deep learning to both the speaker and listener agents, utilizing the advantage of big datasets and massive computation power. As with many other NLP tasks, deep learning has been shown to achieve state-of-the-art results in REG. 
For example, \cite{kazemzadehreferitgame} applied supervised learning and computer vision techniques to referring expression.  
Nevertheless, explainability remains a problem as it is difficult to fully understand how a deep learning model can generate some texts given an image and a target.

On the other hand, recent developments in computational pragmatics have yielded probabilistic models that follow simple conversational rules with great explanatory power. One important example is the Rational Speech Act framework (RSA) by \cite{Frank12predictingpragmatic}, where probabilistic speakers and listeners recursively reason about each other's mental states to communicate---speakers reason about probability distribution over utterances given a referent object, while listeners reason about probability distribution over objects in the scene given an utterance. While \cite{Frank12predictingpragmatic} and \cite[i.a.]{degenRedundancyIsUseful} have shown that RSA can generate sentences that are pragmatically appropriate, the datasets are small, with simple examples that are carefully crafted with perfect information. \cite{andreas-klein-2016-reasoning} extend RSA to real world examples of reference games by using simple shallow models as building blocks to build the speaker and listener agents. \cite{andreas-klein-2016-reasoning}'s approach is intractable though, as the speaker model has to consider all possible utterances. 

\cite{reubenCharacterLevel} resolves this issue by using a character level LSTM to predict one character at a time, thus reducing the search space. 
At each step, instead of generating one utterance, the speaker model generates one character. This method greatly reduces the search space and make the neural RSA system more efficient with harder examples. However, their method is applied to the task of generating a referring expression for an image given several other images instead of a referring expression for an object in a scene. In addition, by performing RSA on a character level \cite{reubenCharacterLevel} partially compromises the explainability of RSA as it is harder to reasoning why at each step, the model would prefer one character over another on describing the target.

To extend on the work of \cite{andreas-klein-2016-reasoning} and \cite{Frank12predictingpragmatic} we want to explore a different approach from \cite{reubenCharacterLevel} that would not compromise on the explainability of RSA. In this paper, we introduce a novel way to apply the RSA framework to real world images and a large scale dataset. Specifically, our contributions are as follows:
\begin{compactenum}
  \item We tackle the intractability problem that \cite{andreas-klein-2016-reasoning} faced, we use Graph R-CNN \cite{jianweiGraphRCNN} and Detectron2 \cite{wu2019detectron2} to extract textual information about objects and their properties i.e., types and attributes, and relations to other objects. This step vastly reduces the search space when generating utterances.
  \item We use the world view generated from the previous step to constrain the utterance space; we also sequentially update the utterance prior and the prior over objects in the scene as each descriptor in the utterance is produced.
  To our knowledge, this is the first attempt to use iterative update of the both the utterance and object probability distributions.
  \item We evaluate our framework on \mbox{refCOCO} and \mbox{refCOCO+} \cite{LichengYuRefCOCO}, and evaluate generated expressions in terms of accuracy (with human evaluation)---whether the expressions are distinctive---and automatic metrics.
  \item We provide a detailed analysis of the result, specifically on the types of error based on the human evaluation. This deviates from standard evaluation process where they key metric is the comprehension accuracy (i.e is the expression distinctively describe the target) and provides a new angle in analysing expression quality.
\end{compactenum}

\section{Background}

\subsection{RSA}

RSA, first introduced by \cite{Frank12predictingpragmatic} encapsulates the idea that pragmatic reasoning is essentially Bayesian. 
In the reference game scenario studied by  \cite{Frank12predictingpragmatic}, the domain consists of a set of objects with various qualities that are fully available to two players. The \textit{speaker} will describe one targeted object unknown to the \textit{listener} by creating a referring expression and the \textit{listener} needs to reason about which object the expression is referring to. As laid out by \cite{scontrasProblang}, RSA is a simple Bayesian inference model with three components: \textit{literal listener}, \textit{pragmatic speaker} and \textit{pragmatic listener}. For a given object $o$ and utterance $u$:

\small
\begin{equation}
     \textit{literal listener}\ P_{L_0}(o|u) \propto \llbracket u\rrbracket(o) \cdot P(o)
\end{equation}
\begin{equation}
    \textit{pragmatic speaker}\ P_{S_1}(u|o) \propto \alpha U(u,o)
\end{equation}
\begin{equation}
     \textit{pragmatic listener}\ P_{L_1}(o|u) \propto P_{S_1}(u|o) \cdot P(o)
\end{equation}
\normalsize

\noindent
where $\llbracket u\rrbracket$ is the literal meaning of $u$, either true (1) or false (0). The \textit{literal listener} thus interprets an utterance at face value, modulo the prior probability of referring to that object $P(o)$, which we take to correspond to the object's salience. The \textit{pragmatic speaker} decides which utterance to make by using the utility function $U(u,o)$, which is a combination of literal listener score and a cost function and the $\alpha$ term denotes the rationality scale of the speaker. Lastly, the \textit{pragmatic listener} infers the targeted object by estimating the likelihood that the \textit{speaker} would use the given utterance to describe it. \cite{Frank12predictingpragmatic} showed that RSA can accurately model human listener behavior for one-word utterances in controlled contexts with few objects and few relevant properties. Since then, a wealth of evidence has accumulated in support of the framework; see \cite{scontrasProblang} for some examples. Still, most RSA models use a very constrained utterance space, each utterance being a single lexical item. 
\cite{degenRedundancyIsUseful} explore RSA models with two-word utterances where each utterance is associated with its own (continuous) semantics. But it remains a major open question how to scale up RSA models for large-scale natural language processing tasks.

\begin{figure*}[hbt!]
    \centering
  \includegraphics[width=0.9\textwidth]{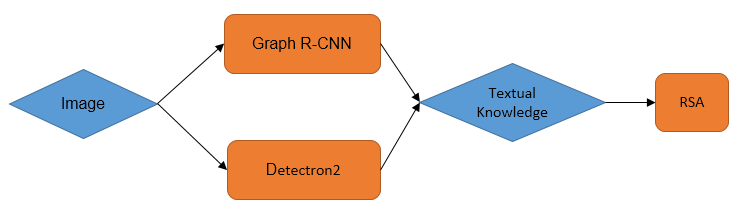}
  \caption{The workflow of Iterative RSA where the input image is passed through information extraction algorithm (Graph R-CNN/Detectron) and pre-processed before given to Iterative RSA.}
  \label{outline}
  \vspace{-1em}
\end{figure*}
\subsection{Detectron2 and Graph R-CNN}
The RSA framework requires prior knowledge about the images and targets in order to generate expressions. Most approaches that use RSA and the speaker/listener model acquire this knowledge through a deep learning model that learns an embedding of the image and the target object, represented as a bounded box or bounded area inscribed on the image then use these embeddings to generate expressions. Instead of using embeddings, we decided to take a different route by generating the symbolic knowledge in the form of scene graph obtained from the image using Detectron2 and Graph R-CNN, which contains objects, properties, and relations, all in a lingual format, which is the ideal input for an RSA model.

Detectron2 is the state-of-the-art object detection model developed by \cite{wu2019detectron2} that utilizes multiple deep learning architecture such as Faster-RCNN \cite{renNIPS15fasterrcnn} and Mask-RCNN \cite{matterport_maskrcnn_2017} and is applicable to multiple object detection tasks. Graph R-CNN \cite{jianweiGraphRCNN} is a scene graph generation model capable of detecting objects in images as well as relations between them using a graph convolutional neural network inspired by Faster-RCNN with a relation proposal network (RPN). RPN and Graph R-CNN is among the state-of-the-art architecture in objects' relation detection and scene graph generation.

\section{Method}
As discussed in \cite{scontrasProblang} and \cite{Frank12predictingpragmatic}, RSA requires a specification of the utterance space and background knowledge about the state of the `world' under consideration. Thus, we view the problem of generating referring expressions as a two-step process where, given an image and a targeted region, we:
\begin{compactenum}
    \item[(1)] Acquire textual classifications (e.g.\ {\em car}) of the objects inside the image and the relations between objects in the image;
    \item[(2)] Generate a referring expression from the knowledge acquired from step (1).
\end{compactenum}

In step (1), most previous work falls into two categories. \cite{Frank12predictingpragmatic} and \cite{degenRedundancyIsUseful} assume the information about objects and their properties are known to the agent generating the expression. On the other hand,  \cite{andreas-klein-2016-reasoning} and \cite{reubenCharacterLevel} use deep learning to obtain embeddings of the image and the targeted region. \cite{RoutianComprehensionGuided} combine the embedding extraction step with the referring expression in one single model. 

In step (1), we neither assume the availability of descriptive knowledge of the images like \cite{Frank12predictingpragmatic} nor do we use an image and region embedding like \cite{andreas-klein-2016-reasoning}. Instead, we generate both the utterance space and the literal semantics of the input image by applying Graph R-CNN to obtain objects' relations and Detectron2 to obtain objects' properties. This idea is motivated by the intractable problem that \cite{andreas-klein-2016-reasoning} face when considering a vast number of utterances at every step. By extracting the symbolic textual information from images, we vastly reduce the number of utterances per step since the number of objects, their relations, and properties are limited in each image. Specifically, Detectron2 outputs objects and the probability that some property is applicable to those objects. For example, a given object categorized as an elephant might have a high probability of having the property \textit{big} and a lower probability of having the property \textit{pink}. Graph R-CNN outputs pairs of objects and probabilities of how true some predefined relation is to some pair of objects.

One challenge in merging computer vision systems with datasets like RefCOCO is matching the target referent in the dataset to the right visually detected object (assuming it is found).
RefCOCO provides a bounding box around the target referent, and Detectron2 and Graph R-CNN may or may not identify an object with the same position and dimensions.
One simple approach is to use the most overlapped detected object with the target box as the subject for the generation algorithm. However, there is no guarantee that the most overlapped detected object is the target. We overcome this problem by combining feature extraction with target feature extraction from Detectron2. We first let Detectron2 identify all the objects it can in the image (call this the {\em context}). We then instruct Detectron2 to consider the target box an object and classify it. If there is an object in the context that overlaps at least 80\% with the target box \textit{and} is assigned the same class, then we leave the context as is; otherwise we add the target box to the context.

\begin{figure*}[hbt!]
\centering
  \includegraphics[width=\textwidth]{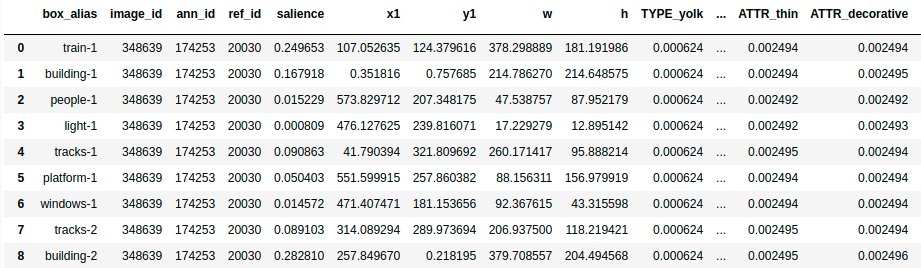}
  \caption{An example of the textual knowledge acquired from Detectron2: each row corresponds to all information about a suggested bounding box, which contains box name, dimension, location, and the likelihoods of types and attributes.}
  \vspace{-1.5em}
  \label{textual_extraction}
\end{figure*}

To enrich object relations beyond binary relations in Graph R-CNN, we also implemented a simple algorithm to generate ordinal relations. We do so by sorting detected objects of the same category (e.g all dogs in an image) by the $x$-axis and assign predefined ordinal relations such as \textit{left}, \textit{right}, or \textit{second from left}. 

The product of these image analysis methods are used in the literal semantics, which 
are categorical, although they are based on the gradient output of Detectron2 and Graph R-CNN, which assigns objects to properties and relations with varying degrees of certainty. 
Since Detectron2 and Graph-RCNN output likelihood values for attributes and types for each object as shown in Figure \ref{textual_extraction}, the last step in the textual extraction process is using a cutoff threshold to decide what level of likelihood make one attribute belongs to a particular object. If the threshold is too low, then objects would contain many irrelevant attributes; if the threshold is too high, there may not be enough attributes to uniquely describe some objects. Currently, we use a hard-coded value that is slightly higher than the minimum value where most of the irrelevant attributes and types are, as examined by hand.

Thus, in the spirit of \cite{lassiter+goodman:2017}, we assume a threshold $\theta$ to decide whether a given type or attribute holds of a given object. Let $F$ be a function that assigns: to each attribute and type, a function from $D$ to [0,1]; and to each relation, a function from $D\times D$ to [0,1], where $D$ is the set of objects in the image. $F$ represents the output of the Detectron2 and Graph R-CNN. For each type, attribute, and relation symbol $u$, $\theta(u)$ is a threshold between 0 and 1 serving as the cutoff for the truthful application of the type, attribute, or relation to the object(s). Then $\llbracket u\rrbracket(o) = 1$ iff $F(u)(o) \geq \theta(u)$, etc. Ultimately we plan to learn these thresholds from referring expression training datasets such as RefCOCO. Currently,  they are fixed by hand: one uniform threshold for types/attributes and relations, respectively. Using categorical semantics rather than the gradient semantics that would be obtained directly from the Detectron2 avoids the well-known problems of modification in fuzzy semantics, a proper solution to which would require conditional probabilities that are unknown \cite{edgington:2001}.

Our key contribution with respect to step (2) is at the speaker level. We introduce {\em iterative RSA}, described in the Algorithm \ref{algo} below. Iterative RSA takes as input the domain of all objects $D$, a prior $P(d)$ over all objects $d\in D$, the referent object $o$ and list of possible `utterances' $U$. Although an utterance may consist of multiple words, each `utterance' here is a single predicate (e.g.\ {\em dog}, {\em second from left}, {\em wearing black polo}). We will use the word `descriptor' instead of `utterance' in this setting, because the strings in question may be combined into a single output that the speaker pronounces once (a single utterance, in the proper sense of the word). Again, we take the prior over objects to be proportional to salience (which we define as object size). Our RSA speaker will iteratively generate one descriptor at a time and update the listener's prior over objects at every step until either (i) the entropy of the probability distribution over objects reaches some desirable threshold $K$, signifying that the listener has enough information to differentiate $o$ among objects in $D$, or (ii) the maximum utterance length $T$ has been reached.

\begin{algorithm}
    \small
    \SetAlgoLined
    \SetKwInOut{Input}{input}
    \SetKwInOut{Output}{output}
    \Input{$o$, $D$, $U$, $P_{D}^0$}
     initialization: $E=[]$ \;
     \While{$t < T$ \& \text{Entropy}$(P_{D}^{t-1}) < K$}{
          $u$  = sample(Speaker $P_{S_1}(u|o,P_{D}^{t-1},U_E)$)\;
           $P_{D}^t$ = Literal listener $P_{L_0}(o|u, P_{D}^{t-1})$\;
          \text{add} $u$ to $E$\;
     }
     
     \Output{$E$}
     \caption{Iterative RSA}
     \label{algo}
\end{algorithm}

In standard RSA, the utility function $U(u,o)$ is defined as $U = \log (P_{L_0}(o|u)) + \text{cost}(u)$ \cite{scontrasProblang}. We define ours as:
\begin{equation}
    U_E = \log (P_{L_0}(o|u) + P_{ngram}(u|E)) + \text{cost}(u)
\end{equation}

where $P_{ngram}$ is the probability of $u$ following the previous $n$ words in $E$. Specifically, we use a 3-gram LSTM model ($n$=3). Figure \ref{outline} outlines our overall workflow.

\section{Experiment and Result}
The framework is implemented in Python and will be made publicly available. In the implementation of Algorithm \ref{algo}, we set $T=4$. This value for maximum utterances per expressions come from the average length of the expressions from our target dataset, both RefCOCO and RefCOCO+ have average length less than $4$ utterances per expression. We evaluate our framework on the test set of RefCOCO and RefCOCO+ datasets released by \cite{LichengYuRefCOCO}. For these two datasets, each data point consists of one image, one  bounding box for a referent (the {\em target box}) and some referring expressions for the referent.
We used pre-trained weights from the COCO dataset for Graph R-CNN and Detectron2. Additionally, we experiment separately with finetuning Detectron on RefCOCO referring expressions. Finally, we test the framework with RefCOCO \textit{Google} split test set and RefCOCO+ \textit{UNC} split test set. 

We evaluate the generated expressions on the test dataset with both automatic overlap-based metrics (BLEU, ROUGE and METEOR) and accuracy (human evaluation) (Table~\ref{perform}). Specifically, we run human evaluation through crowdsourcing site \href{https://prolific.co/}{Prolific} on the following scheme: our IterativeRSA, RecurrentRSA \cite{reubenCharacterLevel} and SLR \cite{yu2017joint} trained on $0.1\%, 1\%$ and $10\%$ of the training sets of RefCOCO and RefCOCO+. For each scheme, we collected survey results for $1000$ randomly selected instance from the RefCOCO test dataset from 20 participants and $3000$ instances from RefCOCO+ test dataset from 60 participants. Each image is preprocessed by adding $6$ bounding boxes on some objects in the image, one of which is the true target. The boxes are chosen from $5$ random objects detected by Detectron2 an the true target object. Each participant is asked to find the matching object given expression for $50$ images through multiple choice questions. In addition, we also manually insert $5$ extra instances where the answer is fairly obvious and use those instances as a sanity check. Data from participants who failed more than half of the sanity checks (i.e $3/5$) was not included in the analysis. Since our referring expressions are generated based on extracted textual information about individual objects and not the raw image as a whole, there are cases where Detectron2 does not recognize the object in the target box or the suggested bounding box from Detectron2 is different in size compared to the target box. In such cases, our algorithm ended up generating an expression for a different observable object than the targeted one. To understand the different types of errors our model makes, we also included additional options in cases where the testers cannot identify a box that matched the expression. Specifically, we added three categories of error when no (unique) matching object is identified:
\begin{compactenum}
    \item nothing in the picture matches the description
    \item several things match this description equally well
    \item the thing that matches the description best is not highlighted
\end{compactenum}

Despite the simplicity of our proposed method, it achieves comparable performance in terms of METEOR score to the Speaker-Listener-Reinforcer(SLR) \cite{yu2017joint}. More importantly, our method outperforms SLR in human comprehension under low training data scheme and RecurrentRSA with both RefCOCO and RefCOCO+.

\begin{table}[thbp!]
\centering
\small
\begin{adjustbox}{center}
\begin{tabular}{lllllll}
\hline \textbf{} &\textbf{True} & \textbf{False} & \textbf{Under-informative}  & \textbf{no-match} & \textbf{not-highlighted} & \textbf{adjusted-accuracy}\\ \hline
IterativeRSA & 27.25 & 13.03 & 11.59 & 44.49 & 3.64 & 52.54 \\
Iterative RSA + f-Det2 & 26.52 & 15.1 & 13.79 & 38.51 & 6.07 & 47.86\\
\hline
SLR-10\% \cite{yu2017joint} & 26.95 & 15.71 & 15.98 & 36.51 & 4.85 & 45.96\\
SLR-1\% \cite{yu2017joint} & 14.3 & 16.24 & 11.96 & 51.13 & 6.37 & 33.65\\
SLR-0.1\% \cite{yu2017joint} & 6.1 & 18.14 & 10.48 & 59.73 & 5.55 & 17.57\\
\hline

\hline
\end{tabular}
\end{adjustbox}
\caption{\label{font-table} Human evaluation response on RefCOCO+ images across IterativeRSA and SLR trained under restricted data.
}
\label{human_eval_types}
\end{table}

 Beside raw accuracy, we also report the accuracy rate using the formula $ adjusted-accuracy = True/(True+False+Underinformative)$ where \textit{Underinformative} counts instances where the expressions correctly refer to the referent objects but are not distinctive enough. Our human evaluation accuracy is slightly less than that of MMI \cite{LichengYuRefCOCO} and while our METEOR score is higher. However, our performance measures fall short when compared to the state-of-the-art extensively trained end-to-end deep neural network model by SLR \cite{kim-etal-2020-conan}. This is to be expected as our method was not trained and does not require training on the specific task of referring expression generation or comprehension. Further performance analysis will be given in the next sections.

\begin{table}
\small
\centering
\begin{tabular}{llll}
\hline \textbf{} & \textbf{bleu} & \textbf{rouge}  & \textbf{meteor}\\ \hline
MMI \cite{LichengYuRefCOCO} & 0.37 &0.333 & 0.136 \\
SLR\cite{kim-etal-2020-conan} & 0.38 & 0.386 & 0.16\\
rerank \cite{RoutianComprehensionGuided} & 0.366 & 0.354 & 0.15 \\
\hline
Iterative RSA & 0.18  & 0.125 & 0.11\\

\hline
\end{tabular}
\caption{ NLP metric comparisons to some previous approaches on RefCOCO+ dataset. 
}
\label{perform}
\end{table}

\begin{table}[thbp!]
\small
\centering
\begin{tabular}{ll|l}
\hline  & \textbf{RefCOCO} & \textbf{RefCOCO+} \\
\hline
Iterative RSA & 28.05 & \textbf{27.25} \\
Iterative RSA + f-Det2 & 41.3 & 26.52 \\
\hline
Recurrent RSA\cite{reubenCharacterLevel} & 26.9 & - \\
SLR-10\% \cite{yu2017joint} & \textbf{66.2} & 26.95 \\
SLR-1\% \cite{yu2017joint} & 49.85  & 14.3 \\
SLR-0.1\% \cite{yu2017joint} & 38.5  & 6.1 \\
\hline
\hline
\end{tabular}
\caption{Raw accuracy of referring expression comprehension evaluated human evaluation on Iterative RSA, Iterative RSA with finetuned Detectron2 (f-Det2), Recurrent RSA and SLR trained with limited data of $0.1 \%, 1\%$ and $10\%$ of RefCOCO and refCOCO+ training set. 
}
\label{human_eval}
\end{table}

\section{Comparison with Recurrent RSA and SLR trained with limited data}
As discussed above, to see the advantages and drawbacks of Iterative RSA, we run human evaluation on generated expressions from RefCOCO and RefCOCO+ datasets and compare Iterative RSA with RecurrentRSA-another RSA approach as well as SLR. From Table \ref{human_eval}, Iterative RSA outperforms RecurrentRSA with $28\%$ compared to $26.9\%$. On the other hand, to make a fair comparison with a deep learning end-to-end approach like SLR, we decided to train SLR with limited training data as Iterative RSA does not require any direct training process. From Table \ref{human_eval}, the Iterative RSA (no training) outperforms all SLR models trained with $0.1\%, 1\%$ and $10\%$ training data  for refCOCO+ dataset and outperform SLR model trained with highly limited training data ($0.1\%$) on RefCOCO. Furthermore, when examining the SLR-generated expressions, we observed that for the model trained and tested on RefCOCO dataset, a lot of the expressions contains positional property of objects such as \textit{left, right}, which makes identifying the target easier when the expression is low quality and incomplete (as a result of training on limited data). Thus, we can see that SLR performs better on RefCOCO than RefCOCO+. On the other hand, IterativeRSA performs more consistently, especially when used without any training or observation of the data. Finetuning the Detectron2 model for object detection with RefCOCO expressions improve the performance on the corresponding dataset, however, using the same model on the RefCOCO+ dataset does not show any significant change in accuracy. 
\begin{figure}[hbt!]
\centering
  \includegraphics[width=0.3\textwidth]{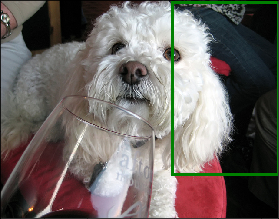}
  \caption{IterativeRSA: {\em jeans}, SLR-$1\%$: \textit{man in black}, SLR-$10\%$: \textit{woman in red}}
  \vspace{-1.5em}
  \label{refcoco+comparison}
  \vspace{1em}
\end{figure}
Figure \ref{refcoco+comparison} is an example of referring expression generated with RSA compared to SLR trained with limited data. For the RSA expression, it clearly shows that the model explains Gricean maxim of quantity by generating the shortest possible word to describe the target which are the jeans, whereas SLR shows the overfitting behavior when generating unrelated expression to the target.
\section{Analysis of the human evaluation}
As mentioned above, in our study, aside from letting users choose one of the objects surrounded by bounding boxes given the generated expression, we also give additional options to handle the case where survey participants cannot find a sensible object to match the description. Overall, we observe that incorrect responses can be  divided into the following categories: \textit{under-informative expression}, \textit{not highlighted}, \textit{no match} and \textit{false}.
These categories of error help in identifying the sources of deficiency in our approach. If the expression is under-informative, 
there are two possibilities. The first is that the textual data extraction step (i.e., Detectron2) was able to identify multiple objects of the same type, but the algorithm is unable to differentiate between the target and the rest of the objects. In this case the problem is on the linguistic side of our model. Another possibility is that not all objects of the relevant type were detected, which is the deficiency of our visual system (Detectron2). 

Another type of visual system deficiency happens when the described object is not the highlighted one or if there is no match. 
In these cases, the visual system (Detectron2) mis-classified the object in the bounding box. As shown in Table \ref{human_eval}, about $48\%$ of the recorded instances belong to these two categories.

\subsection{Under-informative expressions}
\begin{figure}[hbt!]
\centering
  \includegraphics[width=0.3\textwidth]{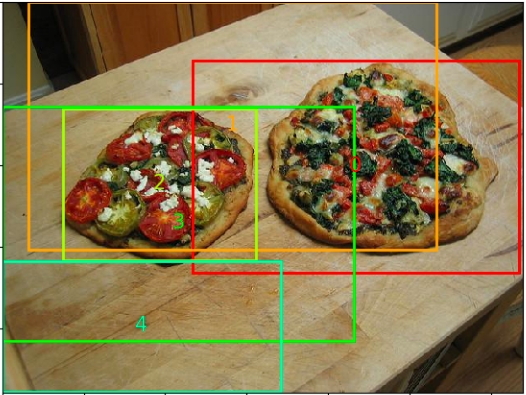}
  \caption{Generated Expression: {\em cooking pizza}, Gold Label: \textit{pizza on left}, Target box: \textit{2}. i.e., the light green box surrounding the smaller pizza.}
  \vspace{-1.5em}
  \label{under_informative}
  \vspace{1em}
\end{figure}

One type of error is when the generated expression is under-informative. This occurs when the expression correctly indicated the type of the target object but failed to differentiate between the target and other objects of the same type in the picture. For example, in Figure \ref{under_informative}, the algorithm was able to correctly identify the type of object in the bounding box but the modifier ({\em cooking}) failed to differentiate the target from the other instance of that type.
\subsection{Object not highlighted}
Another type of errors revealed through human evaluation is when 
the matching object is not highlighted as the target. 
\begin{figure}[hbt!]
\centering
  \includegraphics[width=0.3\textwidth]{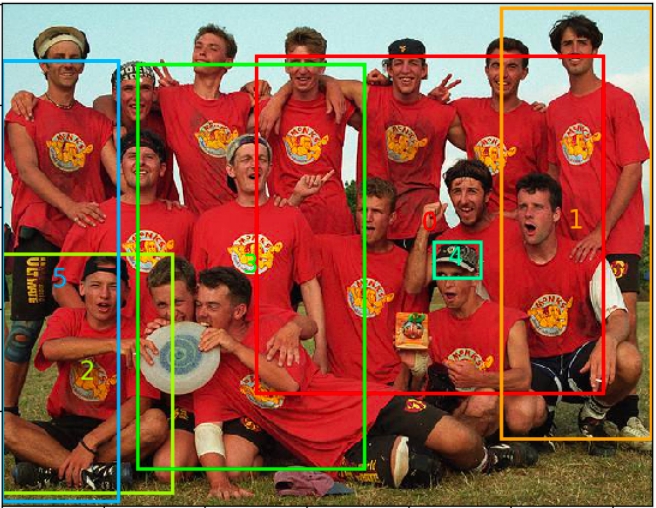}
  \caption{Example of \textit{not highlighted} response Generated Expression: {\em laying down man}, Gold Label: \textit{guy bottom left}, Target box: \textit{2} i.e., the light green box at the bottom left of the image}
  \vspace{-1.5em}
  \label{not_highlighted}
\end{figure}
This type of deficiency is due to the textual extraction component (Detectron2) not observing all objects of the same type. In Figure \ref{not_highlighted}, Detectron2 can only observe four instances of the category \textit{man}, which are all highlighted in this image with box $1,2,3,5$. When comparing the available attributes for these \textit{man}s, target \textit{man} in box 2 (i.e., the light green box at the bottom left of the image) is assigned a distinctive attribute that others do not have: \textit{laying down} (although he is sitting, not laying down). The use of this modifier increases the salience of the target relative to the other individuals that are detected. It is quite possible that participants assumed \textit{laying down man}  refers to the only person at the bottom center of the image who is actually laying down. However, that individual is not detected by Detectron2 and thus there is no highlighted box.

\subsection{High quality expression}
\begin{figure}[hbt!]
\centering
  \includegraphics[width=0.3\textwidth]{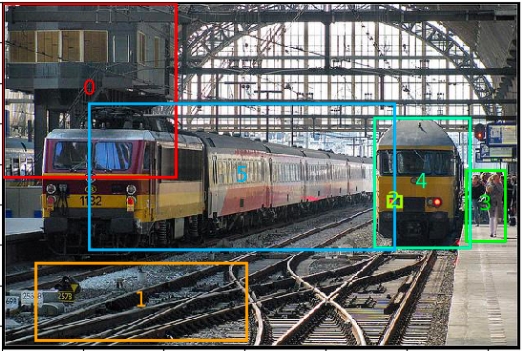}
  \caption{Generated Expression: {\em the right train}, Gold Label: \textit{right train}, Target box: \textit{4}.}
  \vspace{-1.5em}
  \label{correct}
\end{figure}

When the participants correctly identify the target object by choosing the right bounding box, we observe that the textual extraction step provides sufficient information for the algorithm to work correctly. Figure~\ref{correct} is an example where we observe that the system works well when the extracted textual information is accurate and sufficient. Specifically, Detectron2 found all the objects of the type \textit{train} in box $4$ and $5$. Furthermore, the \textit{train} objects have fairly sensible attributes, including \textit{the left} and \textit{the right}.

\section{Discussion}
The Iterative RSA introduced in this paper is able to generate multiple-modifier descriptions, which goes far beyond the vanilla RSA speaker described by \cite{scontrasProblang} and \cite{Frank12predictingpragmatic}, and our RSA speaker has even gone past the two-word stage of \cite{degenRedundancyIsUseful}. While the result is not at the level of the state-of-the-art end-to-end model, Iterative RSA outperforms Recurrent RSA and SLR trained under limited data. We can clearly explain how our model comes up with the referring expressions it generates. The explainability of our model is a contrast feature when compare with RecurrentRSA. While RecurrentRSA also applies the RSA model to generate expressions, its expression generation by recursively generate characters makes it hard to explain why at each step, why one character is a feasible choice that helps identify a target object. Furthermore, to our knowledge, we are the first attempt to apply pure probabilistic RSA model without any neural network components in the expression generation step of the referring expression generation from image task. From the analysis of the human evaluation and concrete examples, it is clear that the performance of Iterative RSA is tightly coupled with the performance of the textual extraction model, particularly Detectron2. When Detectron2 detects enough information, including the objects in a given image as well as their probable attributes, we observe that our proposed Iterative RSA can create high quality expressions with distinctive modifiers. 

Another key strength and also a weakness of our proposed iterative RSA is the size of the vocabulary of descriptors. Currently, this vocabulary is limited to the attributes and types vocabulary that Detectron2 possesses. While this vastly reduces the search space of all possible descriptors, it also limits the possible descriptors that RSA can choose from, given a target. The textual extraction step (Detectron2 in this case) can be analogized to the act of ``observing'' and the Iterative RSA algorithm to ``reasoning''. One cannot reason about objects or aspects of objects that are not observed.

On the other hand, in terms of efficiency, our proposed method is fast because Iterative RSA does not require training data and can be applied directly on the fly with any given textual extraction system. In addition, our application of Detectron2 and Graph-RCNN also does not require training as it utilizes pre-trained weights. Experiments with fine-tuning Detectron2 with RefCOCO data does show better accuracy on the test set of RefCOCO dataset but does not show any major improvement when tested on RefCOCO+ as shown in Table \ref{human_eval_types}. Thus, the base Iterative RSA is more generalized and consistent across different datasets.

Minimal reliance on training data has other advantages: That property makes our approach a promising one for low-resource languages where labeled data for training, especially for vision-language tasks such as referring expression generation/comprehension, are virtually non-existent \cite{joshi-etal-2020-state} for languages other than English. 


\section{Conclusion}
In this paper, we have explored the possibility of decomposing referring expression generation into a two-component process of symbolic knowledge acquisition and expression generation, adapting the RSA framework to real world scenes where textual information is not available. We also introduce two promising innovations that help to address the intractability problem of applying RSA to real world scenes in previous work, which includes (1) constraining the utterance space using the output of object recognition and scene graph generation systems, and (2) proposing a simple yet intuitive and explainable model for referring expression generation called iterative RSA, which incrementally outputs referring expression one predicate at a time. Lastly, our method allows for easy analysis and understanding of each individual expression, and provides clear explanations as to why the system generates the expressions it does.

\bibliographystyle{ieeetr}
\bibliography{ijpla}

\end{document}